\documentclass[letterpaper]{article} 
\usepackage{aaai25}  
\usepackage{times}  
\usepackage{helvet}  
\usepackage{courier}  
\usepackage[hyphens]{url}  
\usepackage{graphicx} 
\usepackage{natbib}  
\usepackage{caption} 
\usepackage{algorithm}
\usepackage{algorithmic}
\usepackage{color}
\usepackage{amssymb}
\usepackage{multirow}
\usepackage{mathrsfs}
\usepackage{colortbl}
\usepackage[table]{xcolor}
\usepackage{amsmath}
\usepackage{bbding}

\usepackage{newfloat}
\usepackage{listings}
\DeclareCaptionStyle{ruled}{labelfont=normalfont,labelsep=colon,strut=off} 
\floatstyle{ruled}
\newfloat{listing}{tb}{lst}{}
\floatname{listing}{Listing}
\title{REGNav: Room Expert Guided Image-Goal Navigation}
\author{
    Pengna Li\equalcontrib, Kangyi Wu\equalcontrib, Jingwen Fu, Sanping Zhou\thanks{Corresponding author.}\\
}
\affiliations{
Institute of Artificial Intelligence and Robotics, Xi’an Jiaotong University\\

    \{sauerfisch, wukangyi747600, fu1371252069\}@stu.xjtu.edu.cn, spzhou@xjtu.edu.cn
}

\usepackage{bibentry}

\begin{document}

\maketitle

\begin{abstract}
Image-goal navigation aims to steer an agent towards the goal location specified by an image. Most prior methods tackle this task by learning a navigation policy, which extracts visual features of goal and observation images, compares their similarity and predicts actions. However, if the agent is in a different room from the goal image, it's extremely challenging to identify their similarity and infer the likely goal location, which may result in the agent wandering around. Intuitively, when humans carry out this task, they may roughly compare the current observation with the goal image, having an approximate concept of whether they are in the same room before executing the actions. Inspired by this intuition, we try to imitate human behaviour and propose a Room Expert Guided Image-Goal Navigation model~(REGNav) to equip the agent with the ability to analyze whether goal and observation images are taken in the same room. Specifically, we first pre-train a room expert with an unsupervised learning technique on the self-collected unlabelled room images. The expert can extract the hidden room style information of goal and observation images and predict their relationship about whether they belong to the same room. In addition, two different fusion approaches are explored to efficiently guide the agent navigation with the room relation knowledge. Extensive experiments show that our REGNav surpasses prior state-of-the-art works on three popular benchmarks.
\end{abstract}

%

\begin{figure}[t]
\includegraphics[width=\linewidth]{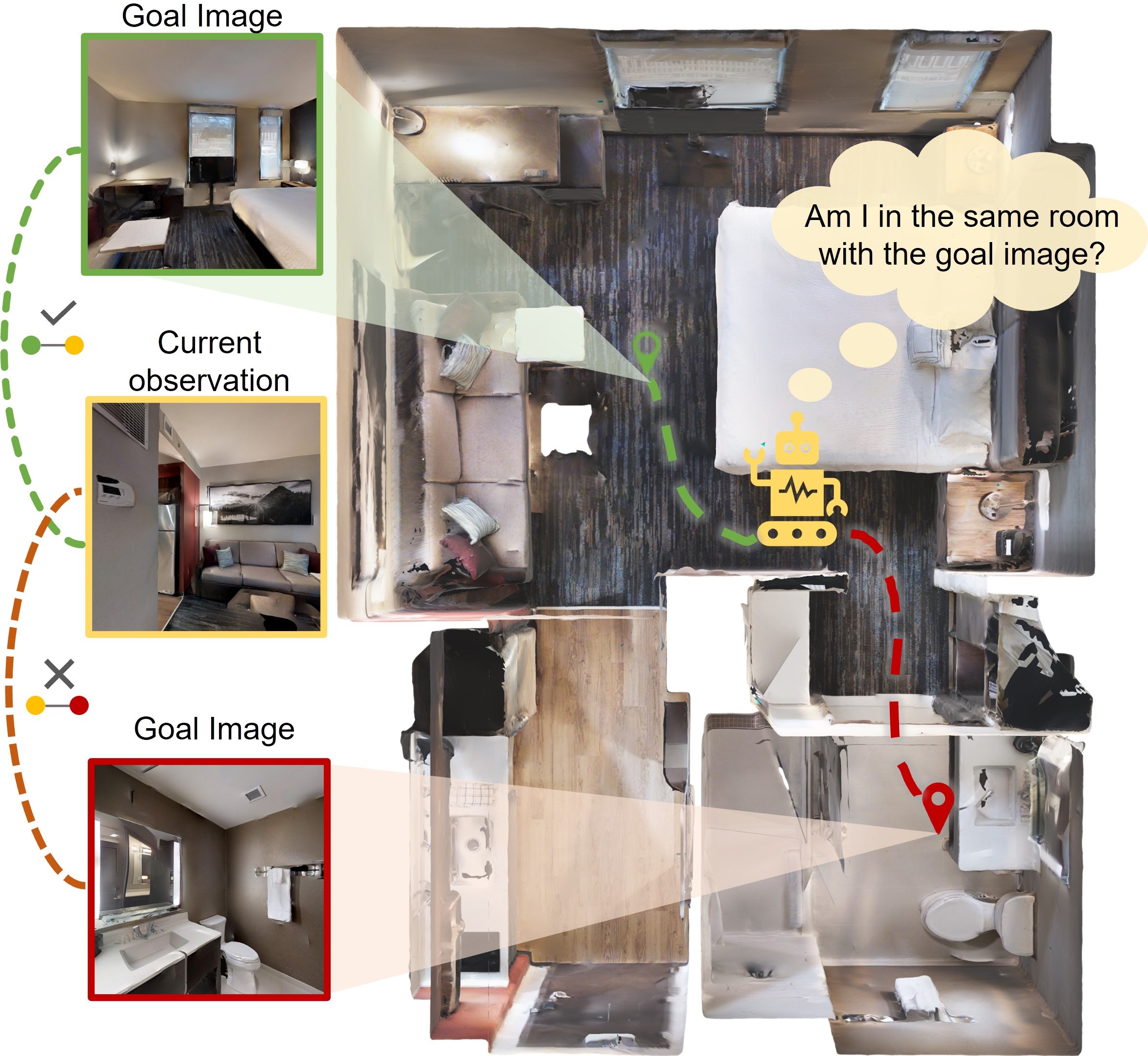}
  \caption{We solve the task of image-goal navigation, where an agent~(the yellow robot) is required to navigate to a location depicted by a goal image. To accomplish this, our agent tries to compare the current observation with the goal image and tease out whether the current location is in the same room with the goal image before executing actions. }
  \label{figure_motivation}
\end{figure}

\begin{links}
    \link{Code}{https://github.com/leeBooMla/REGNav}
\end{links}

\section{Introduction}
Image-goal navigation~(ImageNav)~\cite{zhu2017target} is an emerging embodied intelligence task, where the agent is placed in an unseen environment and needs to navigate to an image-specified goal location using visual observations. 
Due to its widespread applications in last mile delivery, household robots, and personal robots~\cite{wasserman2023last,majumdar2022zson,krantz2023navigating}, it has raised increasing research attention in recent years.

Despite its broad applications, this task remains highly challenging. 
Since the environment map is unknown, the agent must reason the likely location of the goal image to navigate to. This requires the agent to perceive the environment efficiently, compare the current observation with the goal image and find the associations before taking the action. 
However, the complex spatial structure of unseen environments often leads to significant discrepancies between the agent’s actual location and the goal location~(\textit{e.g.} in different rooms). In such cases, the goal image and the current observation may have little overlap and it becomes challenging for the agent to identify their similarities and associations.
This results in the agent failing to reason the goal location, thus taking meaningless actions, such as back-tracking or aimlessly wandering. 
The key to solving this issue is to extract the spatial information from the observations to help reason the spatial relationship with the goal image.

To incorporate spatial information into robot operations, the modular methods~\cite{hahn2021no,chaplot2020neural,krantz2023navigating, lei2024instance} employ GPS or depth sensors to construct a geometric or occupancy map and localize the agent~(SLAM~\cite{durrant2006simultaneous}) in an unfamiliar environment. 
However, these methods heavily rely on depth sensors or GPS to provide spatial information, limiting their scalability in real-world deployment.
Alternatively, the learning-based methods~\cite{yadav2023offline,sun2024fgprompt} attempt to learn an end-to-end navigation policy solely relying on the RGB sensors. These methods directly extract representations of the goal and visual observation to predict the corresponding action. Although the learning-based methods have shown great potential in this task, they have difficulty in exploiting the spatial relationship between the goal and current observation with only RGB sensors available, limiting their performance. 
Imagine that when humans are given the task of finding a place depicted by an image in an unseen environment as illustrated in Figure~\ref{figure_motivation}, 
to find the shortest path to the goal location, humans always try to coarsely estimate spatial relationships about whether the current location is in the same room as the goal image first before further comparing their fine-grained semantics. If not, humans tend to find the door to get out of the current room, which can reduce meaningless actions and move to the target more quickly. Inspired by this, we want to imitate human behaviour and enable the agent to have the ability first to evaluate the coarse spatial relationships between the goal and visual observation, thereby mitigating the issue of the agent’s invalid actions.

To devise such a solution, we have to figure out \textit{what we can rely on to infer the spatial relationships with only RGB images available.} 
As shown in Figure~\ref{figure_motivation}, we observe that different rooms within a house, such as a bedroom, bathroom, and kitchen, often have their specific styles, \textit{e.g.}, decoration style, furniture, floor, and wall. These variations are primarily due to the different functions and requirements of each room. For instance, bedrooms tend to prioritize comfort and they might contain soft lighting, warm colours, and carpeting. Bathrooms often have tiles on the wall and floor for waterproofing and cleaning. 
This observation suggests that it's possible for the agent to identify the room style information from the visual signals. The style information can be used to determine whether the current observation is located in the same room as the goal image.

Learning a model to extract the style information from observation images requires a large amount of annotated image data. However, acquiring the supervision signals is costly and may raise fairness concerns. To address this challenge, we attempt to train a Room Expert with an unsupervised learning method to identify the hidden room style information. Specifically, we utilize the unsupervised clustering with \textit{must-link} and \textit{cannot-link} constraints to pre-train a room-style encoder and a room relation network based on the intuition that if two points are far apart, they are likely located in different rooms.
From the pre-trained model, the agent can obtain the room style representation of the goal and visual observation and obtain their relation about whether they belong to the same room. 

In this paper, we propose a Room Expert Guided Image-Goal Navigation model (REGNav) to explicitly empower the agent with the ability to analyze the spatial relationships between the goal and observation images through a pre-trained room expert model. Specifically, we adopt a two-stage learning scheme: 1) pre-train a room-style expert offline, and 2) incorporate the ability of the pre-trained room expert to learn an efficient navigation policy. The room expert pre-training stage involves adopting an unsupervised learning technique to train a style encoder and a relation network on a large-scale self-collected dataset of images from the indoor environment dataset Gibson~\cite{xiazamirhe2018gibsonenv}. The collected training images share the same parameters and settings as the observations captured by the agent camera. In the latter stage, we explore two different fusion approaches to efficiently guide the agent navigation with the room relation knowledge. We freeze the parameters of the room expert and train the visual encoder and the navigation policy in the Habitat simulator~\cite{savva2019habitat}. Extensive experiments demonstrate that our proposed method can achieve more successful navigation.

We conclude the main contributions of our paper below:

\begin{itemize}
     \item We discuss the issue of the agent's wandering around and explore the feasibility of reasoning the spatial relationships from the pure RGB images.
    \item We observe that room-style information can be the link between the visual signals and spatial relationships. A novel unsupervised method with \textit{must-link} and \textit{cannot-link} constraints is devised to pre-train a room expert to extract room style and predict the spatial relationships.
    \item Finally, We present REGNav, an efficient image-goal navigation framework, equipping the agent with the ability to reason spatial relationships. 
\end{itemize}

\begin{figure*}[t!]
\includegraphics[width=\linewidth]{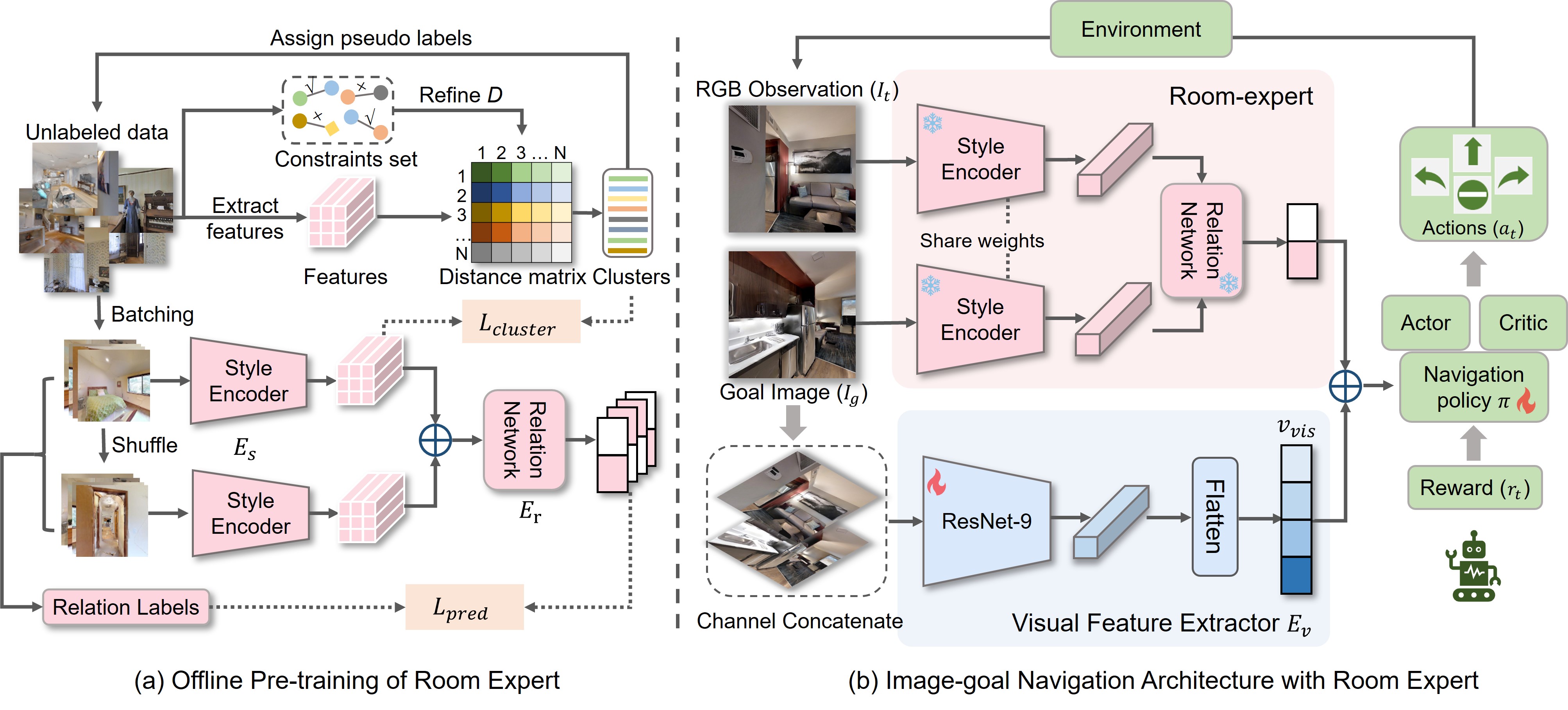}
  \caption{The overview of our REGNav. (a) Pre-training the Room Expert offline. We employ an unsupervised clustering method to train a style encoder and a relation network to extract style representation and predict the relationships. We use the constraints set deduced from the unlabeled data to refine the feature distance matrix to obtain more reliable pseudo labels. (b) The image-goal navigation architecture with Room Expert. We lock the Room Expert and proceed to train the visual encoder and navigation policy. The visual feature extractor regards the channel concatenation of the observation and goal image as input. The navigation policy takes the concatenation of the relation flag~(2-dimension) and the fused feature as input.} \label{figure_framework}
\end{figure*}

\section{Related Works}
\textbf{Visual navigation.} Visual navigation~\cite{krantz2023navigating,kwon2023renderable,pelluri2024transformers,li2023improving,liu2024caven,sun2024fgprompt,wang2024lookahead,zhao2024over} requires an agent to navigate based on visual sensors. It can be categorized into several types, including Visual-and-Language navigation~(VLN), Object Navigation, Image Goal Navigation, etc. Some works~\cite{anderson2018vision,chen2021history,li2022envedit,krantz2023iterative} focus on VLN, which uses additional natural language instructions to depict the navigation targets. These works either depend on detailed language instructions~\cite{wang2023scaling, li2023kerm} or require conversations with humans~\cite{zhang2024dialoc, thomason2020vision} during the navigation process, leading to low usability. 
Object Navigation is proposed with a given object category as the target~\cite{chaplot2020object,mayo2021visual,du2023object, zhang2024imagine}. However, this kind of method can only reach the surrounding area of an object and cannot accurately arrive at a specific location.
Given the reasons above, we address the Image Goal Navigation task where an arbitrary image is provided as the target and only an RGB sensor is utilized during the navigation process. The agent must reach the location depicted in the goal image. We study how to make full use of the knowledge in observation to improve navigation performance.

\noindent\textbf{Reinforcement learning in visual navigation.} Since the image navigation method with reinforcement learning~(RL) can learn directly from interacting with the environment in an end-to-end manner, it has gained a great population in recent years~\cite{du2021curious, majumdar2022zson}. Some methods aim to enhance the representation capability of the feature extractors before the RL policy. \cite{sun2024fgprompt} explore fusion methods to guide the observation encoder to focus on goal-relevant regions. \cite{sun2025prioritized} propose a prioritized semantic learning method to improve the agents' semantic ability. Some works~\cite{li2023improving,wang2024lookahead} utilized the pre-training strategy to enforce the agent to have an expectation of the future environments. 
However, if the agent has a large distance from the goal, these methods may fail to extract useful knowledge from the observations. 
Some methods try to incorporate additional memory mechanisms to enable long-term reasoning and exploit supplementary knowledge from previous states. \cite{mezghan2022memory} trained a state-embedding network to take advantage of the history with external memory. \cite{qiao2022hop} devised a history-and-order pre-training paradigm to exploit past observations and support future prediction. \cite{kim2023topological} inserted semantic information into topological graph memory to obtain a thorough description of history states. \cite{li2024memonav} classified history states into three types to ensure both diversity and long-term memory. However, these methods have no spatial awareness if the agent has never been to the area near the target. On the contrary, we aim to equip the agent with spatial awareness and enable it to analyze whether the observation is in the same space as the goal.

\noindent\textbf{Auxiliary knowledge in visual navigation.} Image-Goal Navigation requires the agent to navigate to an image-specified goal location in an unseen environment using visual observations. Only depending on a single RGB sensor has raised the challenge and makes the task difficult even for humans~\cite{paul2022avlen}. To release the difficulty, auxiliary knowledge is introduced. \cite{liu2024caven} enables the agent to interact with a human for help when it's unable to solve the task. \cite{li2023kerm} utilizes an external pre-trained image description model to provide additional knowledge. \cite{kim2023topological} introduces a pre-trained semantic segmentation model to extract objects in both observations and targets. All of these methods require external datasets to acquire auxiliary knowledge, which may have fairness concerns. In contrast, we devised a Room Expert trained with an unsupervised clustering method without using any additional dataset. Our Room Expert effectively empowers the agent with spatial awareness to analyze the spatial relationship between observations and the goal location, improving the navigation performance.

\section{Proposed Method}
\textbf{Task setup.} 
ImageNav tasks involve directing an agent to a destination depicted in a target image $I_g$, taken at the goal location. Initially positioned at a random starting point $p_0$, the agent is equipped solely with this goal image $I_g$ from the environment. At each time step $t$, it perceives the environment through an egocentric RGB image $I_t$, captured by an onboard RGB sensor. 
Then the agent takes an action conditioned on $v_t$ and $v_g$. These actions denoted as $a_t$, are guided by a trained policy in reinforcement learning frameworks. 

\noindent\textbf{Reward} After acting $a_t$, a reward $r_t$ is given to the agent, encouraging it to reach the goal location through the shortest path~\cite{al2022zero}. Both the reduced distance and the reduced angle in radians are utilized to provide the reward to the agent. The overall reward function for time step $t$ can be formulated as:
\begin{equation}
    r_t = r_d(d_t, d_{t-1}) + \mathscr{I}(d_t \leq d_s)r_\alpha(\alpha_t, \alpha_{t-1}) - \gamma,
\end{equation}
where $r_d$ and $r_\alpha$ are the reduced distance to the goal from the current position and the reduced angle in radians to the goal view from the current view respectively relative to the previous one, $\gamma$ represents a slack reward to encourage efficiency and $\mathscr{I}$ denotes an indicator. What's more, the agent receives a maximum success reward $R_s$
if it reaches the goal and stops within a distance $d_s$ from
the goal location and an angle $\alpha_s$ from the goal angle. The success reward can be formulated as:
\begin{equation}
    R_s = 5 \times [\mathscr{I}(d_t \leq d_s) + \mathscr{I}(d_t \leq d_s\ and \ \alpha_t \leq \alpha_s)],
\end{equation}
Here, we set the $d_s=1$m and $\alpha_s=25^{\circ}$.

\subsection{REGNav}
In this section, we detail the REGNav methodology. It adopts a two-stage learning strategy: 1) pre-training room expert in an unsupervised manner. We first collect a new room-relation image dataset from the indoor dataset Gibson~\cite{xiazamirhe2018gibsonenv}. Based on this, we learn a room expert composed of a style encoder and a relation network in a novel clustering method with constraints set. 2) learning a navigation policy conditioned room expert. We design and explore two different fusion manners to train the visual encoder and navigation policy with the room expert frozen. 

\subsection{Room Expert Pre-Training}
\noindent\textbf{Dataset collection.} 
To obtain the room style representation from observation and goal images, our Room Expert needs to be trained with images taken in different rooms. To avoid focusing on the varied objects between rooms instead of the room style to analyze room relation, images should also be taken from different angles in the same room~(since these images will have completely different objects while they still represent the same room). To ensure the generalizability of the room style representation, images taken in different scenes or houses should also be available. Currently, no publicly available dataset meets the aforementioned requirements. MP3D dataset~\cite{chang2017matterport3d} has provided room annotations. However, previous image-goal navigation models~\cite{sun2024fgprompt, majumdar2022zson} only use Gibson training episode~\cite{mezghan2022memory} to train the agent and use MP3D testing episode~\cite{al2022zero} to evaluate. Directly using room annotations from the MP3D dataset will cause fairness concerns. Therefore, images are collected from the training episodes of Gibson~\cite{mezghan2022memory}
. 
Specifically, for a given training episode $E_m$, we first extract the starting location $p_{ms}$ and the target location $p_{mt}$. Then, agents equipped with a single egocentric RGB sensor are put in these two locations and take images from varied angles. Lastly, we annotate the collected images$\{I_i\}_N$ with scene identity$\{S_i\}_N$ which indicates the 3D scene or house, episode identity$\{E_i\}_N$, and episode difficulty$\{Ed_i\}_N$. We observed that some images collected in this way may contain little room-style information(e.g. when the RGB sensor is too close to the wall, the images taken will be completely black or white.). These blank images, if used in the training process, will provide confusing guidance. To discard these blank images, we input the collected images to SAM~\cite{kirillov2023segment} to get object masks for the whole image. A threshold is set as the minimum object number. Those images whose object mask number is smaller than the threshold are regarded as blank images and are discarded from the dataset. In this way, we build a self-collected dataset to support the training process of the Room Expert to get room style representation. More details can be found in the Appendix.


\noindent\textbf{Unsupervised learning with constraints.} 
Due to the lack of room annotation in the Gibson dataset, we devise a Room Expert composed of a room style encoder and a room relation network trained using an unsupervised clustering algorithm with \textit{must-link} and \textit{cannot-link} to exploit the collected dataset and obtain room-style representation. 

We observe that the Gibson training episodes from \cite{mezghan2022memory} have provided the level of difficulty depending on the distance between the start and target locations: \textit{easy}~(1.5-3m), \textit{medium}~(3-5m) and \textit{hard}~(5-10m). Intuitively, if the two locations are far apart(\textit{hard}), they are most likely in different rooms. Based on this intuition, four rules of room relationship between two arbitrary images $I_i$ and $I_j$ are summarized and a distance refine matrix $M$ with size equal to $N\times N$ is pre-built where $N$ is the number of all collected images: 
\begin{itemize}
\item (1) If $S_i \neq S_j$, then $I_i$ and $I_j$ are not in the same room~(\textit{cannot-link}), set $M_{i,j}=-1$; 
\item (2) If the two images are taken at the same location, then $I_i$ and $I_j$ are definitely in the same room~(must-link), set $M_{i,j}=1$; 
\item (3) If $E_i=E_j$ and $Ed_i=Ed_j=\textit{easy}$, then $I_i$ and $I_j$ are probably in the same room, set $M_{i,j}=0.5$;
\item (4) If $E_i=E_j$ and $Ed_i=Ed_j=\textit{hard}$, $I_i$ and $I_j$ are probably in different rooms, set $M_{i,j}=-0.5$.
\end{itemize}

We build the Unsupervised Room Style Representation Learning based on the four rules above. The framework of Room Expert is illustrated in Figure~\ref{figure_framework}~(a).  Generally, a memory dictionary that contains the cluster feature representations is built and the contrastive loss and cross-entropy loss are utilized to train the Room Expert. Specifically, a standard ResNet-50~\cite{he2016deep} pre-trained on ImageNet~\cite{deng2009imagenet} is used as the backbone for the room-style encoder to extract feature vectors of all the room images. Based on these, we calculate the pair-wise distance matrix $D$ between feature vectors. Then we refine the distance matrix through the pre-built matrix $M$ which serves as the constraints set for feature vectors. The refinement process can be defined as follows:
\begin{equation}
    Refined Distance=D-\gamma M,
\end{equation}
where $\gamma$ the refinement hyper-parameter.
Based on the refined distance matrix, we adopt InfoMap~\cite{rosvall2008maps} clustering algorithm to cluster similar features and assign pseudo labels. With the annotations, we could employ contrastive loss for feature encoder optimization. In this paper, we use the cluster-level contrastive loss~\cite{dai2022cluster}, which is formulated by
\begin{equation}
    L_{cluster} = -log{\frac{exp(E_s(I_i)\cdot \phi_+/\tau)}{\sum_{k=1}^K exp(E_s(I_i)\cdot \phi_k/\tau)}},
 \end{equation}
where $E_s$ represents the style encoder. $K$ is the number of cluster representations and $\phi_k$ denotes the cluster centroid defined by the mean feature vectors of each cluster. $\phi_+$ is a cluster centre which shares the same label with $I_i$. 
Two different image features $I_i$ and $I_j$ are concatenated as input to the room relation network $E_r$ to predict their relation about whether they are taken in the same room. We employ the cross-entropy loss as the relation prediction loss for relation network and style encoder training. The relation predict loss is defined by:
\begin{equation}
\begin{split}
    L_{pred} = -\sum_{n=1}^N y_i\cdot log(E_r(E_s(I_i), E_s(I_j)))+\\
    (1-y_i)\cdot log(1-E_r(E_s(I_i), E_s(I_j))),
\end{split}
\end{equation}
where $E_r$ denotes the relation network and $y_i$ is the relation labels. We jointly adopt the contrastive loss and the relation prediction loss for the room expert training. In summary, the overall objective can be formulated as follows
\begin{equation}
L_{total}=L_{cluster} + \omega L_{pred},
\end{equation}
where $\omega$ represents the hyper-parameter used to balance the two losses.



\subsection{Navigation Policy Learning}
We follow FGPrompt-EF~\cite{sun2024fgprompt} to set up only one visual feature encoder. It concatenates the 3-channel RGB observation $I_t$ with the goal image $I_g$ on the channel dimension and takes the concatenated 6-channel image as the input of the visual feature encoder. We formulate the encoder output as follows:
\begin{equation}
    v_{vis} = E_v(I_t \oplus I_g),
\end{equation}
where $E_v$ is the visual feature encoder and $\oplus$ denotes the channel-wise concatenating.

In this section, we train the visual encoder and navigation policy conditioned on the pre-trained room expert. Two different fusion methods are designed and explored to enable the agent with spatial relation awareness. A naive solution to fuse the knowledge from the room expert is to directly fuse the room-style embedding from the room-style encoder with the visual feature $v_{vis}$. We call this \textit{Implicit Fusion}. These fused features are then fed into the navigation policy~$\pi$ to determine the action $a_t$. In this case, the fusion mechanism can be written as:
\begin{equation}
    a_t = \pi(I_{fusion}(v_{vis}, E_s(I_t), E_s(I_g)))
\end{equation}
Implicit fusion manner requires the agent to distinguish the room relation from room-style embeddings. It's more straightforward to directly give the agent the room relation between the observation and target images and this leads to \textit{Explicit Fusion}. Specifically, the room-style embeddings of the observation and target images $E_s(I_t)$, $E_s(I_g)$ are firstly fed into the room relation network $E_r$ to obtain the spatial relation, as illustrated in Figure~\ref{figure_framework}~(b). The agent is trained to take actions considering this spatial relation. This process can be formulated as:
\begin{equation}
    relation(I_g, I_t) = E_r(E_s(I_t), E_s(I_g)),
\end{equation}
\begin{equation}
    a_t = \pi(E_{fusion}(v_{vis}, relation)),
\end{equation}
The explicit fusion manner is more direct for the navigation policy. More details can be found in the Appendix. 

\begin{table*}[t!]
\centering
\setlength{\tabcolsep}{8pt}
\renewcommand\arraystretch{1.08}
\begin{tabular}{c|cccc|cc}
\hline
\textbf{Method}&\textbf{Reference}&\textbf{Backbone}&\textbf{Sensor}&\textbf{Memory}& \textbf{SPL}$\uparrow$&\textbf{SR}$\uparrow$\\
\hline
ZER&CVPR22&ResNet-9&1RGB&\tiny\XSolidBrush&21.6\%&29.2\%\\
ZSON&NIPS22&ResNet-50&1RGB&\tiny\XSolidBrush&28.0\%&36.9\%\\
OVRL&ICLRW23&ResNet-50&1RGB&\tiny\XSolidBrush&27.0\%&54.2\%\\
OVRL-V2&arXiv23&ViT-Base&1RGB&\tiny\XSolidBrush&58.7\%&82.0\%\\
FGPrompt-MF&NeurIPS23&ResNet-9&1RGB&\tiny\XSolidBrush&62.1\%&90.7\%\\
FGPrompt-EF&NeurIPS23&ResNet-9&1RGB&\tiny\XSolidBrush&66.5\%&90.4\%\\
REGNav&This paper&ResNet-9&1RGB&\tiny\XSolidBrush&\textbf{67.1\%}&\textbf{92.9\%}\\
\hline
\end{tabular}
\caption{Comparison with state-of-the-art methods without external memory on Gibson. 1RGB denotes that only the front RGB sensor is available for the agent and the observation type is one RGB image. All results of these methods are obtained from the overall test set on Gibson.}
\label{tab:sota-gibson}
\end{table*}

\begin{table*}[t]
\centering
\setlength{\tabcolsep}{6pt}
\renewcommand\arraystretch{1.1}
\begin{tabular}{c|cccc|cc|cc|cc}
\hline
\multirow{2}{*}{\textbf{Method}}&\multirow{2}{*}{\textbf{Reference}}&\multirow{2}{*}{\textbf{Backbone}}&\multirow{2}{*}{\textbf{Sensor(s)}}&\multirow{2}{*}{\textbf{Memory}}&\multicolumn{2}{|c|}{\textbf{Easy}}&\multicolumn{2}{|c|}{\textbf{Medium}}&\multicolumn{2}{|c}{\textbf{Hard}}\\
\cline{6-11}&&&&&SPL$\uparrow$&SR$\uparrow$&SPL$\uparrow$&SR$\uparrow$&SPL$\uparrow$&SR$\uparrow$\\
\hline
VGM&ICCV21&ResNet-18&4RGB-D&\checkmark&79.6\%&86.1\%&68.2\%&81.2\%&45.6\%&60.9\%\\
TSGM&CoRL22&ResNet-18&4RGB-D&\checkmark&83.5\%&91.1\%&68.1\%&82.0\%&50.0\%&70.3\%\\
Mem-Aug&IROS22&ResNet-18&4RGB&\checkmark&63.0\%&78.0\%&57.0\%&70.0\%&48.0\%&60.0\%\\
MemoNav&CVPR24&ResNet-18&4RGB-D&\checkmark&-&-&-&-&57.9\%&74.7\%\\
REGNav&This paper&ResNet-9&1RGB&\tiny\XSolidBrush&\textbf{71.4\%}&\textbf{97.5\%}&\textbf{69.4\%}&\textbf{95.4\%}&\textbf{59.4\%}&\textbf{87.1\%}\\
\hline
\end{tabular}
\caption{Comparison with state-of-the-art methods using memory on Gibson. 4RGB denotes that the agent takes a panoramic image from 4 RGB sensors as the observation type. 4RGB-D means that depth image can be used as additional input. The results are evaluated on the easy, medium and hard set of Gibson.}
\label{tab:sota-gibson-memo}
\end{table*}


\begin{table}[t]
\centering
\setlength{\tabcolsep}{6.5pt}
\renewcommand\arraystretch{1.1}
\begin{tabular}{c|cc|cc}
\hline
\multirow{2}{*}{\textbf{Method}}&\multicolumn{2}{|c}{\textbf{MP3D}}&\multicolumn{2}{|c}{\textbf{HM3D}}\\
\cline{2-5}&SPL$\uparrow$&SR$\uparrow$&SPL$\uparrow$&SR$\uparrow$\\
\hline
Mem-Aug&3.9\%&6.9\%&1.9\%&3.5\%\\
ZER&10.8\%&14.6\%&6.3\%&9.6\%\\
FGPrompt-MF&44.3\%&75.3\%&38.8\%&73.8\%\\
FGPrompt-EF&48.8\%&75.7\%&42.1\%&\textbf{75.2\%}\\
REGNav&\textbf{50.2\%}&\textbf{78.0\%}&\textbf{44.0\%}&\textbf{75.2\%}\\
\hline
\end{tabular}
\caption{Comparison of cross-domain evaluation on Matterport~3D~(MP3D) and HabitatMatterport~3D~(HM3D). All methods are trained on the Gibson and directly tested on these two unseen datasets without finetuning.}
\label{tab:cross-eval}
\end{table}

\begin{table}[t]
\centering
\renewcommand\arraystretch{1.1}
\tabcolsep=0.1cm
\begin{tabular}{c|cc|c}
\hline
\multirow{2}{*}{\textbf{Models}}&\multicolumn{2}{|c|}{\textbf{Components}}&\multirow{2}{*}{\textbf{MP3D-Accuracy}$\uparrow$}\\
\cline{2-3}&\textit{Clean Data} &\textit{Refine Dist}\\
\hline
RE-1&&&57.0\%+0.0\%\\
RE-2&\checkmark&&57.4\%+0.4\%\\
RE-3&&\checkmark&57.9\%+0.5\%\\
RE-4&\checkmark&\checkmark&\textbf{58.4\%+0.5\%}\\
\hline
\end{tabular}
\caption{Comparison of the room expert~(RE) and its variants. \textit{Clean Data} refers to the dataset cleaning before the model training. \textit{Refine Dist} is using the constraints set to refine the feature distance matrix. }
\label{tab:ablation-room-expert}
\end{table}


\section{Experiments}
\textbf{Dataset and evaluation metric.} We conduct all of the experiments on the Habitat simulator~\cite{savva2019habitat,szot2021habitat}. We train our agent on the Gibson dataset~\cite{xiazamirhe2018gibsonenv} with the dataset split provided by~\cite{mezghan2022memory}. The dataset provides diverse indoor scenes, consisting of 72 training scenes and 14 testing scenes. We test our agent on the Matterport~3D~\cite{chang2017matterport3d} and HabitatMatterport~3D dataset~\cite{ramakrishnan2021habitat} to validate the cross-domain generalization ability of our agent. 

For evaluation, we utilize the Success Rate~(SR) and Success Weighted by Path Length~(SPL)~\cite{anderson2018evaluation}. SPL balances the efficiency and success rate by calculating the weighted sum of the ratio of the shortest navigation path length to the predicted path length. In an episode, the success distance is within 1m and the maximum steps are set to 500.

\noindent\textbf{Implementation details.} We follow the agent setting of ZER~\cite{al2022zero}. The height of agent is set to 1.5m and the radius is 0.1m.  The agent has a single RGB sensor with a $90^{\circ}$ FOV and 128×128 resolution. The action space consists of MOVE\_FORWARD by 0.25m, TURN\_LEFT, TURN\_RIGHT by $30^{\circ}$ and STOP. For the pre-training stage, we use the Adam optimizer with weight decay 5e-4 and batch size 64 to train the style encoder and relation network for 20 epochs. We set the refinement hyper-parameter $\gamma$ as an adaptive parameter. See the Appendix for the detailed calculation. We set the balance parameter $\omega$ equal to 1. 
For navigation learning, we 
train our REGNav for 500M steps on 8×3090 GPUs. Other hyperparameters follow the ZER.

\noindent\textbf{Baseline.} We build our method on FGPrompt-EF~\cite{sun2024fgprompt}, which involves an agent containing a ResNet-9 encoder for extracting visual features and a policy network composed of a 2-layer GRU~\cite{Chung2014EmpiricalEO}.


\begin{table}[t]
\centering
\setlength{\tabcolsep}{8pt}
\renewcommand\arraystretch{1.1}
\resizebox{8cm}{!}{
\begin{tabular}{c|cc|cc}
\hline
\multirow{2}{*}{\textbf{Datasets}}&\multicolumn{2}{|c|}{\textbf{Implicit fusion}}&\multicolumn{2}{|c}{\textbf{Explicit Fusion}}\\
\cline{2-5}&SPL$\uparrow$&SR$\uparrow$&SPL$\uparrow$&SR$\uparrow$\\
\hline
Gibson&47.4\%&77.0\%&\textbf{67.1\%}&\textbf{92.9\%}\\
MP3D&33.2\%&59.9\%&\textbf{50.2\%}&\textbf{78.0\%}\\
HM3D&27.7\%&55.8\%&\textbf{44.0\%}&\textbf{75.2\%}\\
\hline
\end{tabular}}
\caption{Ablation study of Fusion Manners.}
\vspace{-1mm}
\label{tab:ablation-fusion}
\end{table}

\noindent\textbf{Comparison with SOTAs on Gibson.} 
We report the results averaged over 3 random seeds.~(The variances are less than 1e-3). 
As demonstrated in Table~\ref{tab:sota-gibson}, we first compare our proposed methods with recent state-of-the-art image-goal navigation methods without additional external memory, which don't use the agent's depth or pose sensor. These methods includes ZER~\cite{al2022zero}, ZSON~\cite{majumdar2022zson}, OVRL~\cite{yadav2023offline}, OVRL-V2~\cite{yadav2023ovrl} and FGPrompt~\cite{sun2024fgprompt}. Our REGNav shows a promising result with SPL = 67.1\% and SR = 92.9\% on the overall Gibson dataset. 

We also provide several recent memory-based methods for comparison, including VGM~\cite{kwon2021visual}, TSGM~\cite{kim2023topological}, Mem-Aug~\cite{mezghan2022memory} and MemoNav~\cite{li2024memonav}. Mem-Aug categorized the test episodes of Gibson into three levels of difficulty
We evaluate our REGNav on the corresponding set and report the results in Table~\ref{tab:sota-gibson-memo}. our proposed method illustrates superior performance, outperforming the memory-based methods by a large margin, which indicates the capacity of REGNav to effectively leverage the style information.

\begin{figure*}[t!]
\hspace{0.2cm}
\includegraphics[width=0.91\linewidth]{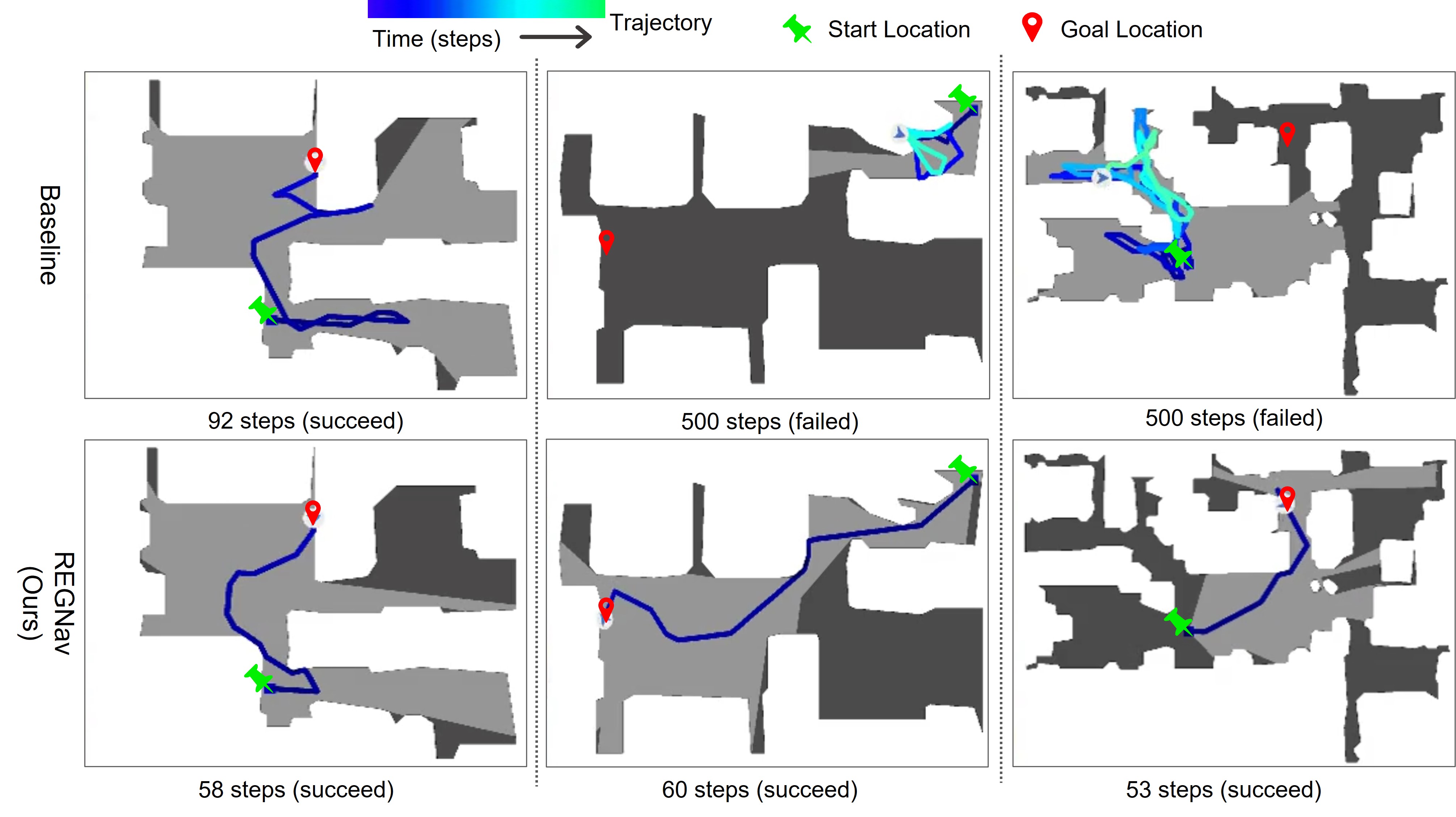}
  \caption{\textbf{The visualization results of example episodes from a top-down view. }The lines originating from the green locations refer to the agent's trajectories, where the colour changes as the steps. The grey regions on the top-down map represent the areas explored by the agent's camera. Compared with the baseline, our REGNav plans more efficient navigation paths.} \label{figure_vis}
\end{figure*}

\noindent\textbf{Cross-domain evaluation.} To prove the domain generalization capability of our REGNav, we evaluate the Gibson-trained models on the Matterport~3D~(MP3D) and HabitatMatterport~3D~(HM3D) without extra finetuning. Cross-domain evaluation is an extremely challenging setting due to the visual domain gap between these datasets. Table~\ref{tab:cross-eval} reports the comparison results. The results of Mem-Aug~\cite{mezghan2022memory} and ZER~\cite{al2022zero} are cited from their paper, while the results of FGPrompt~\cite{sun2024fgprompt} are evaluated from the trained models that FGPrompt released. Compared with previous methods, our REGNav achieves comparable performance on SPL and SR, which shows that focusing on spatial information can lead to better generalization.

\noindent\textbf{Ablation study on Room Expert training scheme.} We investigate
the necessity of cleaning the dataset using SAM~\cite{kirillov2023segment} and refining the feature distance matrix using \textit{must-link} and \textit{cannot-link} constraints set in the pre-training room expert stage. Due to the lack of pair annotations in Gibson, we follow the data collection technique in Gibson to collect a validation dataset in MP3D which has room labels. All the results are trained in the Gibson-collected dataset with unsupervised clustering and evaluated in the MP3D-collected validation set with real labels. We report the relation accuracy of input pair images as the evaluation metric. As shown in Table~\ref{tab:ablation-room-expert}, using both data cleaning and distance refinement are superior to the counterparts, validating the effectiveness of these components.

\noindent\textbf{Comparison of different fusion manners.} We investigate two different fusion methods of incorporating the room-level information of visual observations into the semantic information. Implicit fusion refers to using the room-style embedding of the pre-trained model while explicit fusion is to directly use the room relation between the goal and observation. We report the comparison results in Table~\ref{tab:ablation-fusion} and the more straightforward explicit fusion manner can obtain better performance. This validates that the direct room relation prior can empower the agent with more successful and efficient navigation than the implicit representation.

\noindent\textbf{Visualization.} To qualitatively analyze the effect of our proposed method, we visualize the navigation results using top-down maps. We compare our REGNav with the baseline~(FGPrompt-EF) for different scenes in the Gibson test set in Figure~\ref{figure_vis}. When there exist certain discrepancies between the goal location and start location, due to the lack of spatial relationship priors, the agent of FGPrompt needs to take more steps and frequently wander around, especially in narrow pathways. In contrast, REGNav can analyze the spatial relationships and reason the relative goal location. Therefore, it can efficiently reduce redundant actions and achieve shorter navigation paths, which validates the superiority of REGNav in planning better paths. We also provide more visualization and analysis in the Appendix.

\frenchspacing
\section{Conclusion}
In this paper, we introduced REGNav to address the issue of the agent's meaningless actions for the ImageNav task. Our motivation draws on human navigation strategies, enabling agents to evaluate spatial relationships between goal and observation images through a pre-trained room expert model. This model uses unsupervised learning to extract room style representations, determining whether the current location belongs to the same room as the goal and guiding the navigation process. Our experimental results highlight REGNav's superior performance in planning efficient navigation paths, particularly in complex environments where traditional models struggle with spatial discrepancies.

\section{Acknowledgments}
This work was supported in part by National Natural Science Foundation of China under Grants 62088102 and 12326608, 
Natural Science Foundation of Shaanxi Province under Grant 2022JC-41, Key R$\&$D Program of Shaanxi Province under Grant 2024PT-ZCK-80, and Fundamental Research Funds for the Central Universities under Grant XTR042021005.
\bibliography{aaai25}

\end{document}